\documentclass[conference, compsoc, twoside]{IEEEtran}
\usepackage{fancyhdr}
\usepackage{times}
\usepackage{epsfig}
\usepackage{graphicx}
\usepackage{amsmath}
\usepackage{amsmath}
\usepackage{amsthm}
\usepackage{algorithm} 
\usepackage{algorithmic}
\usepackage{subfigure}
\usepackage{amssymb}
\usepackage{multirow}
\usepackage{multicol}

\ifCLASSOPTIONcompsoc
  \usepackage[nocompress]{cite}
\else
  \usepackage{cite}
\fi
\ifCLASSINFOpdf
\else
\fi
\begin{document}


\title{Robust Emotion Recognition from Low Quality and Low Bit Rate Video: \\ A Deep Learning Approach
\vspace{-1.5em}
}
\author{\textit{Bowen Cheng\textsuperscript{\dag}, Zhangyang Wang\textsuperscript{\ddag}, Zhaobin Zhang$^\Diamond$, Zhu Li$^\Diamond$, Ding Liu\textsuperscript{\dag},} \\
\textit{Jianchao Yang\textsuperscript{\S}, Shuai Huang\textsuperscript{$\flat$}, Thomas S. Huang\textsuperscript{\dag}}
 \\
\dag \,Beckman Institute, University of Illinois at Urbana-Champaign\\
\ddag \,Department of Computer Science and Engineering, Texas A\&M University\\
$\Diamond$ Department of Computer Science \& Electrical Engineering, University of Missouri, Kansas City\\
\S \,Snap Inc, USA $\quad$
$\flat$ Department of Industrial and Systems Engineering, University of Washington\\
{\tt\small \{bcheng9, dingliu2, t-huang1\}@illinois.edu} $\qquad$ {\tt\small atlaswang@tamu.edu}\\
{\tt\small  \{zzktb@mail., lizhu@\}umkc.edu} $\qquad$ {\tt\small jianchao.yang@snap.com} $\qquad$ {\tt\small shuaih@uw.edu}
}


% compsoc conferences compared to traditional conferences), use this
%Telephone: (800) 555--1212, Fax: (888) 555--1212}


\maketitle
\thispagestyle{fancy}
\begin{abstract}
Emotion recognition from facial expressions is tremendously useful, especially when coupled with smart devices and wireless multimedia applications. However, the inadequate network bandwidth often limits the spatial resolution of the transmitted video, which will heavily degrade the recognition reliability. We develop a novel framework to achieve robust emotion recognition from low bit rate video. While video frames are downsampled at the encoder side, the decoder is embedded with a deep network model for joint super-resolution (SR) and recognition. Notably, we propose a novel \textit{max-mix} training strategy, leading to a single ``One-for-All'' model that is remarkably robust to a vast range of downsampling factors. That makes our framework well adapted for the varied bandwidths in real transmission scenarios, without hampering scalability or efficiency. The proposed framework is evaluated on the AVEC 2016 benchmark, and demonstrates significantly improved stand-alone recognition performance, as well as rate-distortion (R-D) performance, than either directly recognizing from LR frames, or separating SR and recognition.
\end{abstract}


% \fi
\IEEEpeerreviewmaketitle

\section{Introduction}
\vspace{-0.5em}
\begin{figure}[tbp]
\centering
\begin{minipage}{0.50\textwidth}
\centering {
\includegraphics[width=\textwidth]{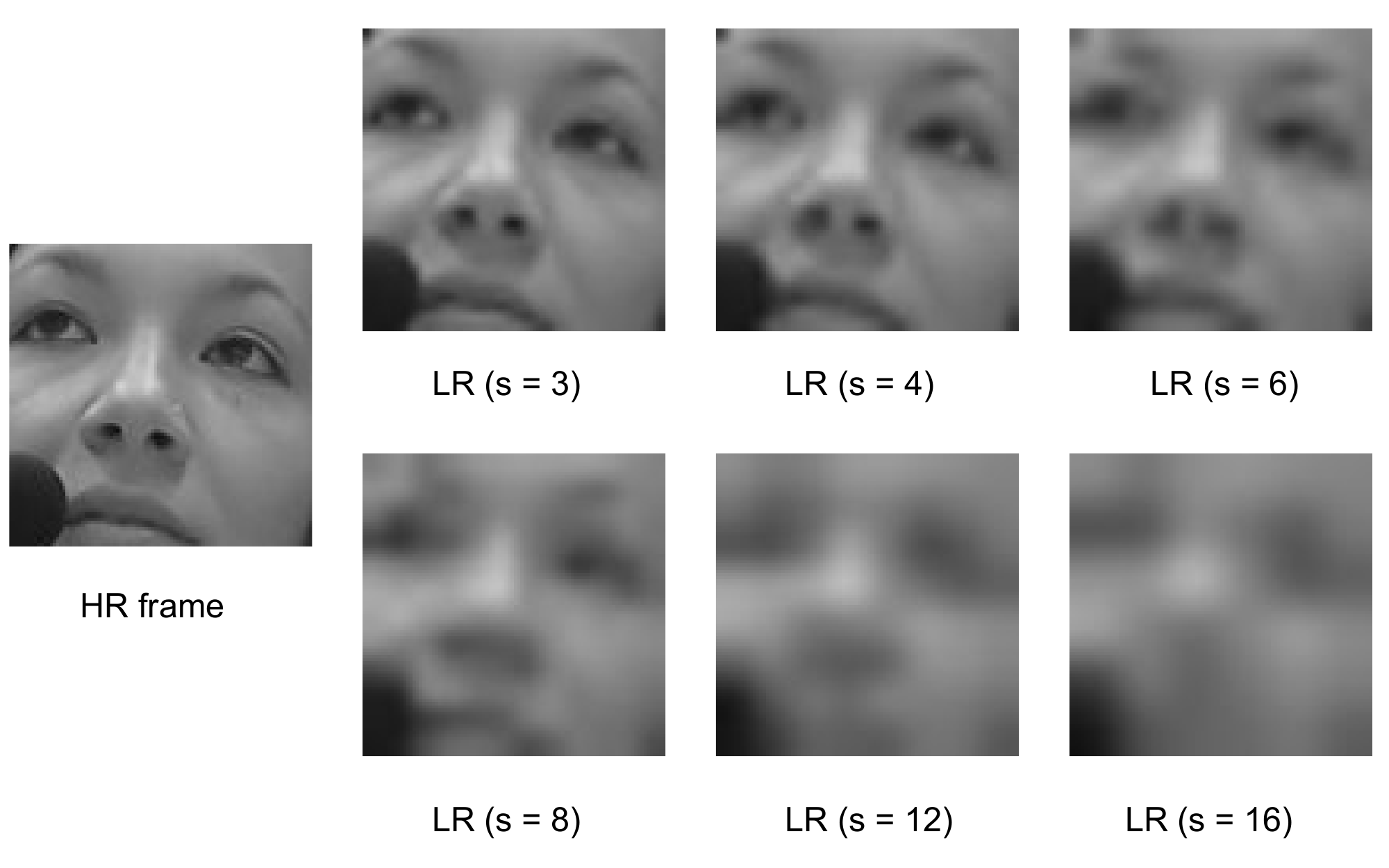}
}\end{minipage}
\caption{A HR face image (resolution: $96 \times 96$) detected from one frame in the dev 8 set of the AVEC 2016 dataset, and its downsampled LR versions with different downsampling factors $s$: \{3, 4, 6, 8, 12, 16\} (displayed after bi-cubic interpolation). Note that our proposed approach can substantially improve the emotion recognition performance, for up to $s$ = 8.}
\label{face}
\end{figure}

Emotion recognition from facial expressions mostly relies on data collected in a highly controlled environment with high resolution (HR) frontal faces. Coupled with the widespread use of smart and wearable devices, emotion recognition techniques have demonstrated the tremendous application value, in tracking human mental status and detecting mental illness, in a less obtrusive way than traditional mental healthcare monitoring approaches \cite{zhou2015tackling}. However, with the ever-growing use of wireless multimedia applications, the available network bandwidth is often inadequate to stream HR video. To transmit video contents over limited bandwidth networks, the encoder often compromises the spatial resolution of video frames for reducing the bit rates, by adaptive downsampling of the HR video to low resolution (LR) prior to compression \cite{nguyen2008adaptive}. It yields improved performance than coding with the original full-size video, yet at the expense of degrading quality. In particular, the LR facial images after decompression constitutes a severe challenge for facial expression analysis \cite{tian2004evaluation}. Figure \ref{face} displays a few examples after downsampling, which apparently make emotion recognition increasingly difficult, or even impossible. 

This paper presents a novel framework to achieve robust and reliable emotion recognition, while keeping the communication load low. At the encoder side, the video frames are adaptively downsampled before compression and transmission, in order to meet the bandwidth requirements. The core innovation of the proposed framework is a jointly optimized scheme of super resolution (SR) and recognition models based on deep learning \cite{krizhevsky2012imagenet}, after decoding. As an important finding, we develop a novel ``max-mix'' training strategy, and obtain a single deep model that is verified to be robust to a vast range of downsampling factors. The ``One-for-All'' model is well adapted for the varied bandwidths in practical transmission. Our model demonstrates significantly superior recognition and rate-distortion (R-D) performance, than either directly recognizing from LR frames, or the two-stage pipeline where restoration and recognition are separate. Finally, we point out a few directions, towards which our framework can be further improved.

\section{Related Work}
\vspace{-0.5em}
\begin{figure*}[htbp]
\centering
\begin{minipage}{0.99\textwidth}
\centering {
\includegraphics[width=\textwidth]{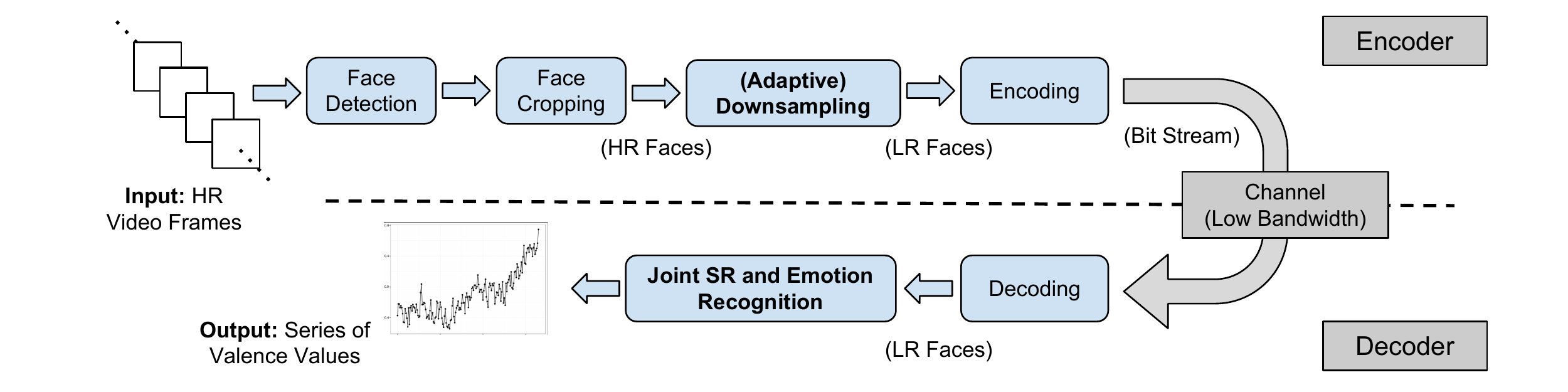}
}\end{minipage}
\caption{Pipeline of the proposed framework. The intermediate outputs are also annotated along with the pipeline.}
\vspace{-0.5em}
\label{pipeline}
\end{figure*}

\subsection{Emotion Recognition from Facial Expressions}
\vspace{-0.5em}
Recognizing human emotion can depend upon gesture, pose, facial expression, speech, behaviors, and even brain signals \cite{wang2017image}. In this paper, we mainly discuss emotion recognition from videos that record facial expressions. The seminal work \cite{tian2001recognizing} recognized fine-grained changes in facial expression by proposing the Facial Action Coding System (FACS). A large portion of research efforts tried to formulate emotion recognition as a multi-class \textit{classification} problem. The most famous categorization system is the scheme of  six ``universal'' atom emotions \cite{krause1987universals}: anger, disgust, fear, happiness, sadness, and surprise. Many feature engineering or feature learning approaches have been proposed for the six-emotion classification problem, e.g., \cite{shan2009facial, kahou2013combining, liu2014facial, khorrami2015deep}. 
%However, such six atom emotions cannot well represent all varied human emotions exhaustively. 

The \textit{regression} formulation is another promising alternative to model the infinite space of possible emotions \cite{russell1977evidence}. A person's emotions were found to be described by a low-dimensional representation. One simple and common choice is to decompose the emotion into two orthogonal and real-valued dimensions: \textit{arousal} and \textit{valence} \cite{kensinger2004remembering}. Arousal measures how engaged or apathetic a subject appears, while valence measures how positive or negative a subject appears. The arousal-valence representation describes a larger and continuous space of emotions, which the six-emotion scheme only roughly partitions the emotion space into six regions. Moreover, the regression formulation allows for time-continuous, real-valued outputs, which is more realistic for modeling temporal emotion dynamics from video. 
%It is thus adopted by many recent works \cite{chao2015long, ebrahimi2015recurrent, icip}.

Several benchmarks have been constructed for the task of automatic emotion recognition, such as the extended Cohn-Kanade (CK+) dataset \cite{lucey2010extended}, and the MMI facial expression database \cite{valstar2010induced}. Following many recent works \cite{chao2015long, ebrahimi2015recurrent, icip}, we develop our emotion recognition model based on the AVEC 2016 \cite{ringeval2015av+} dataset, whose data was originally from the RECOLA corpus \cite{ringeval2013introducing}. Multimodal signals, including audio, video (40 ms binned frames), and physiological signals, were synchronously recorded from 27 subjects. Continuous-time and continuous-valued ratings of arousal and valence were given by human raters. In this paper, we focus on video data only, and choose the valence value as the regression goal for simplicity (same as \cite{icip}, one of the state-of-the-arts on the same dataset). The proposed method can integrate other data modalities, and can be easily extended to predict arousal and valence values jointly. 

\vspace{-0.5em}
\subsection{Low Bit Rate Video Transmission with Adaptive Downsampling}
\vspace{-0.5em}
For a variety of computer vision tasks where processing server needs to communicate with remotely deployed visual sensors, the communication costs can be prohibitive, especially for applications like city-scale visual surveillance networks, where thousands of high resolution cameras are connected. How to reduce the communication cost in the distributed vision system is an important research issue. 

Extensive prior works have shown that downsampling to LR prior to encoding and upsampling after decoding can 
can reduce the operating cost in bit rate, and with upscaling/super-resolution, can visually beat the video compressed directly at HR using standard codecs with the same number of bits, under insufficient bit rates  \cite{bruckstein2003down, lin2006adaptive, nguyen2008adaptive}. In addition, video downsampling has also been a common practice pre-processing for high-level computer vision tasks such as detection and tracking, in order to meet the computational complexity and/or latency requirements, especially on mobile devices with limited processing power \cite{chen2016evaluate}.

At the decoder side, SR techniques are often adopted as post-processing for enhancing the display quality \cite{ma2009block, shen2011down}. If a fixed downsampling ratio during encoding is known, the SR models can be obtained by various example-based training approaches \cite{yang2010image, wang2015learning, wang2015self, liu2016robust}. However, the practical bandwidth might be varied due to network load, congestion and bottleneck situations. \cite{wang2014adaptive, dong2014adaptive} pointed out that to achieve the overall optimal R-D performance, the downsampling ratio at the encoder had to be adaptively determined. In that way, the distortions caused by downsampling which reduces the number of pixels transmitted, and coding which introduces quantization noises to the pixels transmitted, could be balanced. As a result, the SR post-processing at the decoder side has to effectively cope with varied downsampling factors. One straightforward but expensive solution is to utilize an ensemble of SR models, each of which is trained dedicatedly for one downsampling factor. A more cost-effective option is to seek a single ``one-for-all'' SR model, whose performance keeps robust over a useful range of low resolutions. Up to our best knowledge, its viability has not been examined yet.

\vspace{-0.5em}
\subsection{Low-Resolution Visual Recognition}
\vspace{-0.5em}
Empirical studies \cite{lui2009meta, 80tiny} in face recognition proved that a minimum face resolution between $32 \times 32$ and $64 \times 64$ is required for most stand-alone recognition algorithms, whose performance would be much degraded when applied with even lower resolutions \cite{cvpr08, VFR}. In the emotion recognition literature, most existing methods assumed the availability of HR frontal faces. \cite{tian2004evaluation} first investigated the effects of different image resolutions for facial expression analysis. The author concluded that while the performance difference was negligible when the head region resolution was $72 \times 96$ or higher, the recognition turned growingly unreliable when head region resolution was lower than $36 \times 48$. It is thus desirable to obtain more robust features for LR images and low-intensity expressions \cite{shan2009facial}

When dealing with LR subjects, the traditional two-stage pipeline tried to first apply SR algorithms before perform recognition tasks. Recently, the SR performance has been noticeably improved, with the aid of deep network models \cite{Tang}. However, the recovered HR images inevitably over-smoothened details. More importantly, such a straightforward approach yields the sub-optimal performance: the artifacts introduced by the reconstruction process will undermine the final recognition. \cite{zhang2011close} presented a close-the-loop approach of  image restoration and recognition, based on the assumption that the degraded image, if correctly restored, will also have a good identifiability. \cite{vlrr} advanced the methodology using a deep network trained from end to end, and observed the possibility of robust object recognition even when the region of interests (ROI) was smaller than $16 \times 16$ pixels. However, it remains to be an open issue how much the performance degradation can be remedied in the same way for emotion recognition. 

%On the other hand, it  is noteworthy that although HR images are not available in real testing, it could still be utilized in training as auxiliary information to obtained enhanced features.

\section{Technical Approach}
\vspace{-0.5em}
\subsection{System Overview}
\vspace{-0.5em}
The pipeline of the proposed framework is illustrated in Figure \ref{pipeline}. We assume that face detection and cropping has been accomplished at the encoder side as pre-processing. Only the cropped faces are to be downsampled, compressed and transmitted to the decoder side \cite{andalo2016transmitting}. After decoding, the joint SR and emotion recognition module simultaneously enhances the spatial resolution and predicts the per-frame valence value, using an end-to-end deep network, which will be detailed in the next section. The system outputs a time series of predicted valence values.  

We do not discuss how to adaptively control the downsampling factors as per the communication needs, which has been well studied in previous video coding and wireless communication literature \cite{wang2014adaptive, dong2014adaptive}. Instead, we aim to make the decoder robust to a wide range of varied downsampling factors that the encoder might adopt.

\begin{figure}[tbp]
\centering
\begin{minipage}{0.50\textwidth}
\centering {
\includegraphics[width=\textwidth]{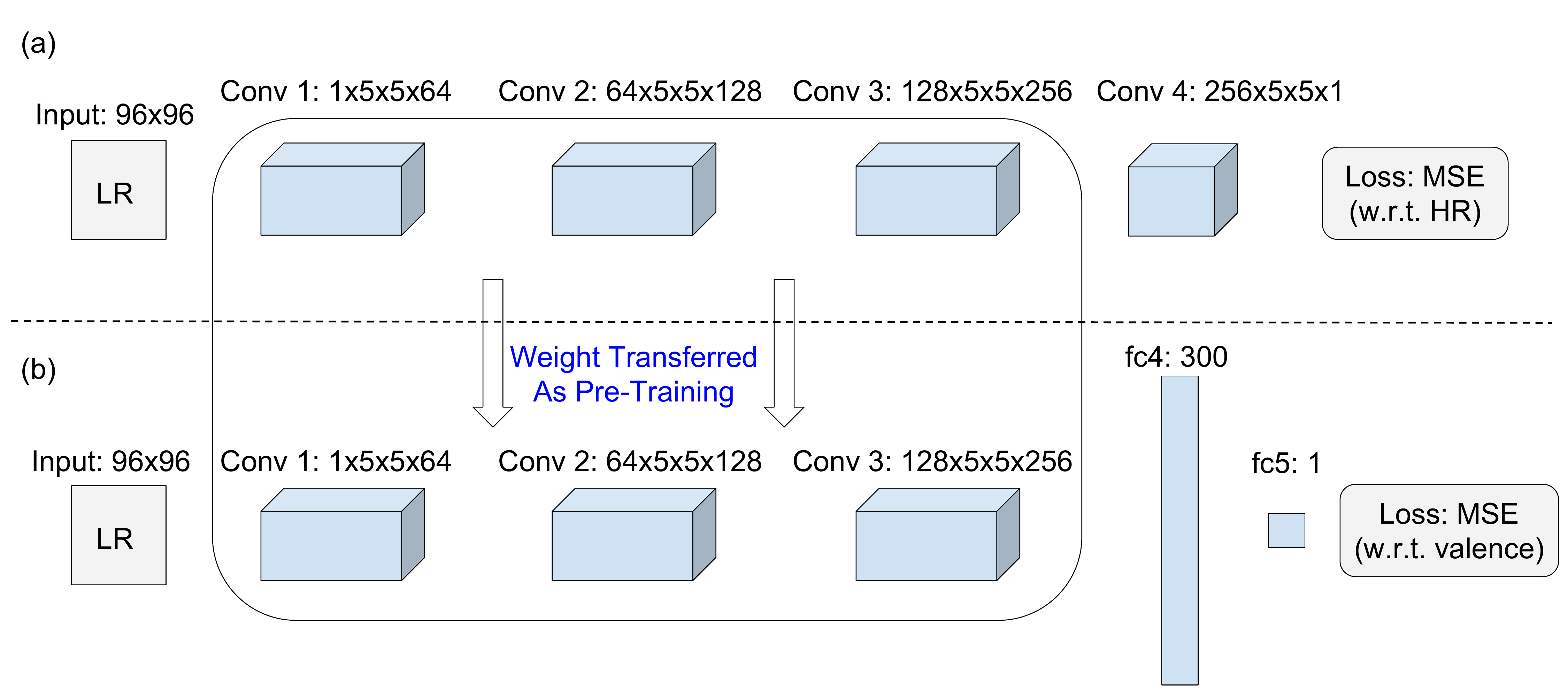}
}\end{minipage}
\caption{The network architectures for: (a) SR fully convolutional network (SR-FCN); (b) Convolutional neural network (CNN) for joint SR and emotion recognition.}
\vspace{-0.5em}
\label{cnn}
\end{figure}

\vspace{-0.5em}
\subsection{Joint SR and Emotion Recognition}
\vspace{-0.5em}
Figure \ref{cnn} (b) depicts the convolutional neural network (CNN) architecture for joint SR and emotion recognition, which mostly inherits the CNN+D structure in \cite{icip}. The target CNN is fed with LR video frames. It has 3 convolutional layers consisting of 64, 128, and 256 filters respectively, each of size $5 \times 5$. The first two layers are followed by $2 \times 2$ max pooling while the third layer is followed by quadrant pooling. Followed is a fully-connected layer with 300 hidden units, regularized by dropout with probability 0.5. ReLU neuron is adopted for all. A linear regression layer estimates the valence values, under the mean squared error (MSE) loss function. 

As pointed out by \cite{vlrr}, training a CNN-based recognition model over LR images is usually not robust and prone to overfitting, due to the severe information loss. On the other hand, a CNN trained on HR images will also witness degraded performance when tested on LR images, due to the domain mismatch. Our main intuition is to regularize and enhance the CNN feature extraction, by pre-training the first several convolutional layers using a SR sub-model, which reconstructs HR images from LR counterparts. 

A 4-layer SR fully convolutional network (SR-FCN) is first constructed, as in Figure \ref{cnn} (a). Its first three layers are configured the same as the first three layers of the target CNN, while the fourth layer reconstructs the input image from the output feature maps of the third layer. SR-FCN is trained in an unsupervised way to reconstruct the HR frames from LR inputs, under the MSE loss as well. Note that it is different from the target CNN that regresses LR frames to valence values. After that, its first three layers are exported to initialize the first layers of the target CNN. Starting from this SR-based partial initialization, the CNN is then jointly tuned for the emotion recognition task, from end to end. 

\begin{table*}[tbp]
%\footnotesize
\begin{center}
\begin{tabular}{|c|c|c|c|c|c||c|c|c|c|}
\hline
\multirow{2}{*}{} & \multirow{2}{*}{HR} & \multicolumn{4}{|c||}{$s$ = 3} & \multicolumn{4}{|c|}{$s$ = 4}  \\ 
\cline{3-10}
 & & LR-3 & Non-Joint-3 & Joint-3 & Joint-OA & LR-4 & Non-Joint-4 & Joint-4 & Joint-OA  \\\hline
RMSE & 0.146 & 0.142 & \textbf{0.121} & 0.132 & 0.129 & 0.155 & \textbf{0.123} & 0.127 & 0.131 \\
\hline
CC & 0.430 & 0.392	& 0.363& 0.396	 & \textbf{0.399} & 0.381& 0.354 & 0.380 & \textbf{0.391}\\
\hline
CCC & 0.325 & 0.302 & 0.293 & 0.323 & \textbf{0.328} & 0.283& 0.281& 0.319 & \textbf{0.327} \\
\hline
\hline
\multirow{2}{*}{} & \multirow{2}{*}{HR} & \multicolumn{4}{|c||}{$s$ = 6} & \multicolumn{4}{|c|}{$s$ = 8} \\ 
\cline{3-10}
 & & LR-6 & Non-Joint-6 & Joint-6 & Joint-OA & LR-8 & Non-Joint-8 & Joint-8 & Joint-OA \\
 \hline
RMSE & 0.146 & 0.149 & 0.128 & \textbf{0.127} & 0.134 & 0.161 & \textbf{0.129} & 0.134 & 0.130\\
\hline
CC & 0.430 & 0.300 & 0.344 & 0.325 & \textbf{0.375} & 0.323 & 0.317 & 0.320 & \textbf{0.358} \\
\hline
CCC & 0.325 & 0.263 & 0.280 & 0.274 & \textbf{0.309} & 0.238 & 0.265 & 0.266 & \textbf{0.285} \\
\hline
\hline
\multirow{2}{*}{} & \multirow{2}{*}{HR} & \multicolumn{4}{|c||}{$s$ = 12} & \multicolumn{4}{|c|}{$s$ = 16} \\ 
\cline{3-10}
 & & LR-12 & Non-Joint-12 & Joint-12 & Joint-OA & LR-16 & Non-Joint-16 & Joint-16 & Joint-OA \\
 \hline
RMSE & 0.146 & 0.143 &0.126 &  \textbf{0.125} & 0.132 & 0.137 & \textbf{0.124} & 0.137& 0.132 \\
\hline
CC & 0.430 & 0.291 & 0.287 & 0.246 & \textbf{0.308} & \textbf{0.316} & 0.244 & 0.219 & 0.273 \\
\hline
CCC & 0.325 & 0.224 & \textbf{0.235} & 0.204 & 0.223 & \textbf{0.212} & 0.191 & 0.192 & 0.172 \\
\hline
\end{tabular}
\end{center}
\caption{The overall RMSE, CC and CCC comparisons at different factors $s$ (best results in each case are in bold).}
\label{table}
\vspace{-2em}
\end{table*}

\vspace{-0.5em}
\subsection{Max-Mix Training for An One-for-All Model} 
\vspace{-0.5em}
Almost all data-driven SR approaches \cite{yang2010image, Tang} as well as some latest low-resolution recognition works \cite{zhang2011close, vlrr} assume one identical downsampling factor between training and testing. A SR model is only dedicated to coping with one downsampling factor. It is more desirable to train a \textit{``One-for-All''} model, since it is robust to the vast range of downsampling factors caused by the varied  transmission bandwidths, without incurring any scalability or efficiency issue. Given a range of possible downsampling factors, we propose the \textit{max-mix} training: first pre-training SR-FCN with LR-HR pairs generated with the \textit{maximum downsampling factor}, followed by fine-tuning the CNN model, on a mixture of LR frames that are generated from HR frames using \textit{the range of all downsampling factors}\footnote{In our experiments, we find that mixing all LR frames of $s$ = [3, 4, 6, 8, 12, 16] does not lead to the optimal performance. We conjecture that``bad'' $s$ values such as 12, 16 lead to un-recognizable LR samples that perturb training. Instead, we mix a ``reasonable'' range of LR samples of $s$ = [3, 4, 6] for fine-tuning. It is the default way to obtain the Joint-OA model, and is verified to be better than fine-tuning with any single $s$.}. As verified by our experiments, the resulting CNN is able to achieve even better performance, than dedicatedly trained SR models at a specific downsampling factor. 

\section{Experiments}
\vspace{-0.5em}
For all AVEC video data, we first convert color frames to gray-scale, and crop the face from each video frame using the given bounding box. All face regions are then normalized to $96 \times 96$ pixels, and are treated as the HR subjects to be downsampled, compressed and transmitted. We generate LR frames using a range of downsampling factors $s$: [3, 4, 6, 8, 12, 16]. Such a range is intentionally set to be vast: while $s$  = 3 causes only mild degradations, $s$ = 16 leads to $6 \times 6$ facial regions whose expressions are unlikely to be identified even by human viewers. 

%Readers may refer to \cite{ringeval2015av+} for a more comprehensive description of the AVEC dataset.

All CNNs were trained using stochastic gradient descent with batch size of 128, momentum of 0.9, and weight decay of $5\times10^{-4}$. We apply mean subtraction and contrast normalization prior to passing each face image through the CNN. We train the SR-SCN for 30,000 iterations, using a constant learning rate of 0.01 is used, and . To fine-tune the target CNN, a learning rate of 0.001 is used for the first three pre-trained layers, and the remaining layers are initialized randomly and trained with a learning rate of 0.01: the learning rates are both divided by 10 when we observe that the validation set performance stops to improve.  

We use the AVEC development set of 9 sequences as our testing set. Three metrics are measured for the emotion recognition performance \cite{ringeval2015av+}: (i) Root Mean Square Error (RMSE); (ii) Pearson Correlation Coefficient (CC); and (iii) Concordance Correlation Coefficient (CCC), which combines CC with the RMSE between the mean of the two compared time series. A good recognition result will likely favor \textit{lower RMSE}, as well as \textit{higher CC and CCC}. Note that CCC is the \textit{most reliable} measure among the three, and was thus used to choose AVEC competition winners.

\begin{figure*}[htbp]
\centering
\begin{minipage}{0.245\textwidth}
\centering \subfigure[$s$ = 3, CC] {
\includegraphics[width=\textwidth]{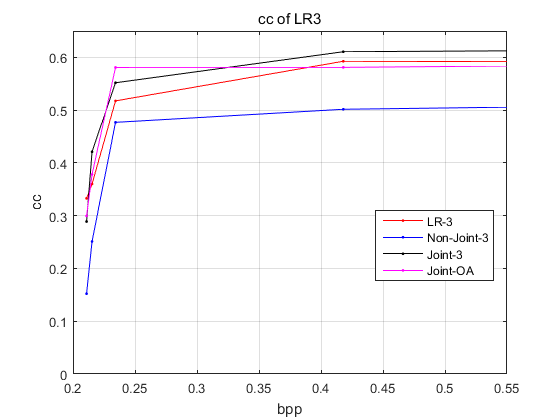}
}\end{minipage}
\begin{minipage}{0.245\textwidth}
\centering \subfigure[$s$ = 3, CCC] {
\includegraphics[width=\textwidth]{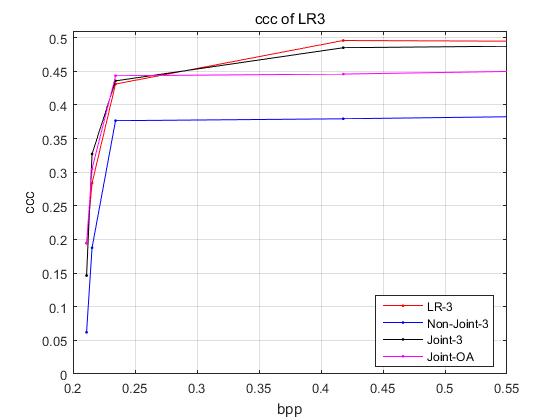}
}\end{minipage}
\begin{minipage}{0.245\textwidth}
\centering \subfigure [$s$ = 4, CC] {
\includegraphics[width=\textwidth]{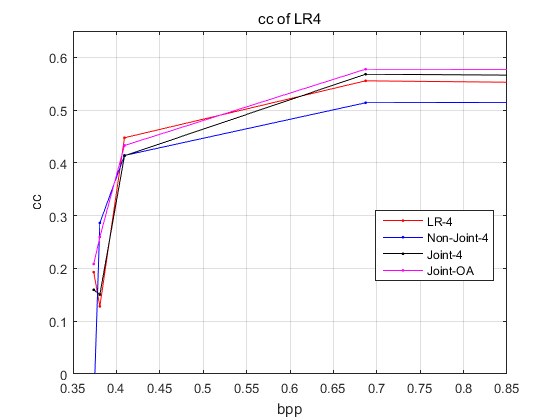}
}\end{minipage}
\begin{minipage}{0.245\textwidth}
\centering \subfigure [$s$ = 4, CCC] {
\includegraphics[width=\textwidth]{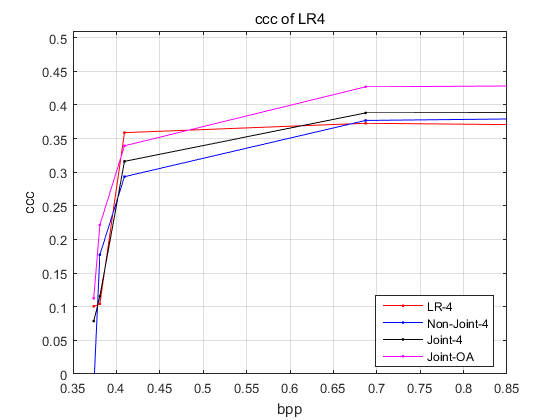}
}\end{minipage}
\caption{The CC and CCC comparisons at different QPs, with $s$ = 3 and 4.}
\vspace{-0.5em}
\label{QP}
\end{figure*}

\vspace{-0.5em}
\subsection{Performance Evaluation and Analysis of Low-Resolution Emotion Recognition}
\vspace{-0.5em}
We consider the following comparison methods:
\begin{itemize}
\vspace{-0.3em}
\item \textbf{HR:} a CNN baseline trained and tested on HR data.
\item \textbf{LR-$s$:} a CNN baseline trained and tested on LR data, with the downsampling factor $s$.
\item \textbf{Non-Joint-$s$:} a SR-FCN is first trained to up-scale LR frames to HR. A separate fully-connected neural network is then trained to regress predict valence values from up-scaled HR images. The SR and emotion recognition modules are not jointly tuned. 
\item \textbf{Joint-$s$:}  the joint SR and emotion recognition model described in Section 3.2, trained dedicatedly for a specific downsampling factor $s$. 
\item \textbf{Joint-OA:}  the joint SR and emotion recognition model, training with the max-mix strategy.  
%Generate $\{\mathbf{y}_i\}$ and $\{\mathbf{z}_i\}$ with downsampling factors $\alpha$ and $\beta$ respectively ($\beta \ge \alpha$), and train $\M$ using Algorithm \ref{dpretrain}. LQ-$\alpha$-$\beta$-joint is reduced to LQ-$\alpha$-joint when $\alpha$ = $\beta$.
\vspace{-0.3em}
\end{itemize}
For fair comparison, we carefully ensure all models to have the same amount of parameters. Table \ref{table} presents the overall RMSE, CC and CCC comparison results on the AVEC development set \footnote{We follow \cite{icip} to first concatenate all nine sequences into one long sequence, and then compute its RMSE/CC/CCC as the overall results.}. Comparison HR and LR-$s$ certifies the notable impact of low resolution on emotion recognition. 

If we look at RMSEs only, then non-joint methods achieve best in almost all cases (even better than HR). However, RMSE results display little consistency with CC/CCCs, implying that RMSE may not be a reliable measure. For $s$ = 12 and 16, little improvement seems attainable over the LR baselines, since all recognizable information are almost lost at such low resolutions (see Fig. \ref{face}). For $s$ = 3, 4, 6 and 8, the CC and CCC results are fairly consistent: the recognition benefits from joint training in most cases. What is more, the Joint-OA model consistently outperforms Joint-$s$, with the largest margins of 0.050 (CC) and 0.035 (CCC) at $s$ = 6. With surprise, we notice that for $s$ = 3, 4, the Joint-OA results even slightly surpass HR in terms of CCC.

Two questions arise naturally: (1) why joint training can help; and (2) why a ``distracted'' Joint-OA model can beat ``dedicated'' Joint-$s$ models? \textit{For Question 1}, the SR hallucinated details help discover subtle features, which are otherwise prone to be overlooked in LR frames \cite{vlrr}. However, the restoration-driven pre-training non-selectively enhances all visual details, which may also include artifacts that hamper recognition. The joint tuning step introduces extra information (the valence values) to reinforce the learning of more task-related features, while suppressing other unrelated components. \textit{For Question 2},  we conjecture that pre-training SR-FCN with maximum $s$ helps its low-level filters to capture more robust mappings, boosting the (implicit) feature enhancement. Further, the mixture $s$ fine-tuning may correspond to re-scaling training data, which is a popular type of data augmentation for classification tasks \cite{krizhevsky2012imagenet} and helps learn scale-invariant features. 

\vspace{-0.5em}
\subsection{Rate-Distortion Performance Comparison}
\vspace{-0.5em}
For bandwidth constrained applications, achieving robust facial expression recognition from low bit rate video can be attractive for many security and surveillance applications. The problem is coupled with the low resolution sensor problem, but has its own peculiar challenges. In addition to the loss of pixels from resolution limitations, video coding may also introduce quantization errors that can affect the emotion recognition performance. Indeed, compression of visual features for visual recognition has been an active research topic with many interesting results for key point feature compressions \cite{nagar2014akula, xin2013laplacian}.

In this experiment, we encode the actively downsampled testing video at different quality-rate levels to mimic real world transmissions, where video is usually coded subject to a rate constraint. We then fed the decoded videos to the joint SR and recognition models (same as Section 4.1 \textit{without re-training}), and calculate the CC and CCC results. The observations in Figure \ref{QP} are mostly consistent with the uncompressed case, showing our models' robustness to coding qualities. For $s$ = 3, Joint-OA gains more advantages with larger quantization parameters (QPs)\footnote{the smaller the QP is, the better the reconstruction quality would be.}
, while for $s$ = 4 Joint-OA outperforms other methods for most bit rates. For the Rate-Distortion (RD) operating range with good to excellent visual quality, the loss of recognition performance is negligible. As coding-introduced distortion becomes more pronounced at larger QPs, the recognition starts to suffer. 

With $s$ = 3, 4, the CC and CCC starts to saturate for QPs smaller than 24, which operate at approximately 0.24 bits per pixel (bpp) for $s$ = 3, and 0.4 bpp for $s$ = 4. The loss of coding efficiency in LR4, compared to LR3, is due to the fixed overhead from video coding headers and structures, that is shared among all pixels. The efficiency decreases as the number of the pixels is reduced. Notice that the pixels fed into the recognition algorithm are 8-bit. This compression is indeed effective on top of the active downsampling in conserving the bandwidth. 
%This is indeed very similar to the Rate-Distortion performance of the sequence.

In summary, actively downsampling reduces the number of pixels to be transmitted, while coding with larger QPs enforces heavier quantization of the pixels remaining. Both will contribute to saving the bandwidth, and there exists an interesting tradeoff in-between.

%Acknowledgement: AVEC, Tom and Pooya
\vspace{-0.5em}
\section{Conclusion and Discussion}
\vspace{-0.5em}
This paper presents a novel framework for robust emotion recognition from low bit rate video, and demonstrates its promising performance as well as strong robustness to both pixel reduction and pixel quantization. There is apparent room for its further performance improvement. From the \textit{system perspective}, we expect to incorporate more building blocks (e.g., the video encoding and decoding steps) into the joint optimization scheme, and make the pipeline in Figure \ref{pipeline} more end-to-end. From the \textit{model perspective}, so far we have not utilized any temporal information for video-based recognition. The previous work \cite{ebrahimi2015recurrent,icip} exploited recurrent neural networks to capture the temporal coherence, and obtained additional performance gains. Since adjusting the \textit{temporal resolution}  (a.k.a., frame rate) \cite{yu2013multi} is also a common means to reduce video bit rates, our future work may also extend to adaptive temporal downsampling, followed by temporal-spatial joint video SR and recognition. Finally, as we observe that CC/CCC are evidently better evaluation metrics than RMSE, it is a noteworthy option to train our emotion recognition model under CC/CCC-based loss functions rather than the current MSE loss.

\vspace{-0.5em}
  \section*{Acknowledgments}
\vspace{-0.5em}
Bowen Cheng, Ding Liu and Thomas Huang's research works are supported in part by US Army Research Office grant \textit{W911NF-15-1-0317}. The authors sincerely acknowledge the valuable efforts of the AVEC challenge organizers \cite{ringeval2015av+}. The authors would also like to acknowledge the helpful discussions with Dr. Pooya Khorrami and Dr. Thomas Paine.

\bibliographystyle{IEEEtran}
\bibliography{references}
% \bibliographystyle{IEEEbib}
% \bibliography{acii}

\end{document}